\documentclass[lettersize,journal]{IEEEtran}
\usepackage{amsmath,amsfonts}
\usepackage{algorithm}
\usepackage{array}
\usepackage{textcomp}
\usepackage{stfloats}
\usepackage{url}
\usepackage{verbatim}
\usepackage{float}
\usepackage{subfigure}
\usepackage{graphicx}
\usepackage{multirow}
\usepackage{cite}

\usepackage{algorithm}
\usepackage{algorithmic}
\usepackage{graphicx}
\usepackage{amsmath}
\usepackage{amssymb}
\usepackage{color}
\usepackage{epstopdf}
\usepackage{array}
\usepackage{multirow}
\usepackage{amsfonts}
\usepackage{booktabs}
\usepackage{float}
\usepackage{subfigure}
\usepackage[switch]{lineno}

\usepackage{xcolor}

\begin{document}

\title{Dynamic Correlation Learning and Regularization for Multi-Label Confidence Calibration}
\author{Tianshui Chen, Weihang Wang, Tao Pu, Jinghui Qin, Zhijing Yang, Jie Liu, Liang Lin \emph{Fellow, IEEE}
\thanks{Tianshui Chen, Weihang Wang, Jinghui Qin, and Zhijing Yang are with the Guangdong University of Technology (Email: tianshuichen@gmail.com, wwkml994@gmail.com, scape1989@gmail.com, yzhj@gdut.edu.cn). Tao Pu and Liang Lin are with the Sun Yat-Sen University (Email: putao537@gmail.com, linliang@ieee.org). Jie Liu is with North China University of Technology (Email: liujxxxy@126.com). (\emph{Corresponding author: Jinghui Qin}). }
\thanks{This work was supported in part by National Natural Science Foundation of China (NSFC) under Grant No. 62206060 and No. 62206314, in part by Guangzhou Basic and Applied Basic Research Foundation under Grant No. SL2022A04J01626 and No. 2022A1515011835, Science and Technology Projects in Guangzhou under Grant No. 2024A04J4388.}
}

\maketitle
\begin{abstract}
Modern visual recognition models often display overconfidence due to their reliance on complex deep neural networks and one-hot target supervision, resulting in unreliable confidence scores that necessitate calibration. While current confidence calibration techniques primarily address single-label scenarios, there is a lack of focus on more practical and generalizable multi-label contexts. This paper introduces the Multi-Label Confidence Calibration (MLCC) task, aiming to provide well-calibrated confidence scores in multi-label scenarios. Unlike single-label images, multi-label images contain multiple objects, leading to semantic confusion and further unreliability in confidence scores. Existing single-label calibration methods, based on label smoothing, fail to account for category correlations, which are crucial for addressing semantic confusion, thereby yielding sub-optimal performance. To overcome these limitations, we propose the Dynamic Correlation Learning and Regularization (DCLR) algorithm, which leverages multi-grained semantic correlations to better model semantic confusion for adaptive regularization. DCLR learns dynamic instance-level and prototype-level similarities specific to each category, using these to measure semantic correlations across different categories. With this understanding, we construct adaptive label vectors that assign higher values to categories with strong correlations, thereby facilitating more effective regularization. We establish an evaluation benchmark, re-implementing several advanced confidence calibration algorithms and applying them to leading multi-label recognition (MLR) models for fair comparison. Through extensive experiments, we demonstrate the superior performance of DCLR over existing methods in providing reliable confidence scores in multi-label scenarios.

\end{abstract}


\section{Introduction}
\IEEEPARstart{M}{odern} visual recognition models, built on complex deep neural networks \cite{he2016deep,dosovitskiy2020image}, often suffer from overfitting to training data, inevitably leading to overly confident and unreliable predicted score dilemma \cite{mukhoti2020calibrating, guo2017calibration,lu2023prediction}. This dilemma severely prevents their applications to high-risk scenarios, such as self-driving \cite{grigorescu2020survey, janai2020computer} and medical diagnosis \cite{jiang2012calibrating, caruana2015intelligible}. To deal with this issue, numerous works \cite{muller2019does, mukhoti2020calibrating, liang2020improved} are intensively proposed for confidence calibration that can provide more accurate and reliable predicted confidence scores to indicate an accurate probability of correctness. Despite achieving impressive progress, these efforts predominantly concentrate on single-label settings, where each image is associated with a single category. However, these works can hardly be applied to multi-label scenarios,  which are more reflective of real-world scenarios where images often contain objects from multiple categories \cite{wei2016hcp,chen2022sst,zhang2023spatial}. Our work targets the multi-label confidence calibration (MLCC) task, seeking to extend and enhance calibration techniques for these more complex and practical scenarios.

\begin{figure}[!t]
  \centering
  \includegraphics[width=0.99\linewidth]{main_pic.pdf}
  \vspace{-10pt}
  \caption{Two examples of predicted scores by current advanced MLR models with and without DCLR calibration. The categories existing in the image are highlighted in \textbf{bold}. }
  \label{fig:mlcc-vis}
\end{figure}

Current multi-label recognition (MLR) models \cite{chen2019learning,chen2019multi} mainly use the one-hot target labels for each class as supervision, overlooking the information about other categories. 
Consequently, these models are prone to indiscriminately assigning overconfident scores to their predictions, culminating in the critical issue of overconfidence. 
Moreover, these models either learn holistic features \cite{chen2019multi,ye2020attention} or category-specific features \cite{chen2019learning} for prediction. However, given the presence of multiple semantic objects scattered throughout an image, these features often capture information from various semantic objects, leading to significant semantic confusion and further exacerbating the overconfidence issue. In the first example shown in Figure \ref{fig:mlcc-vis}, existing models exhibit confusion between ``vase" and ``potted plant" due to their similar appearances, leading to an overly confident score being assigned to the non-existent ``potted plant". Similarly, objects resembling ``cell phone" and ``book" are {also non-existent}, and the models mistakenly allocate high-confidence scores to these two categories. Traditional confidence calibration algorithms \cite{muller2019does,mukhoti2020calibrating} typically employ the label smoothing mechanism \cite{muller2019does}, which smooths each category by equally and independently reallocating a small value to all non-target categories from the target category. However, these algorithms do not consider category correlations that can capture semantic confusion to obtain adaptive regularization, leading to sub-optimal performance in the MLCC scenario.

In this work, we introduce the Dynamic Correlation Learning and Regularization (DCLR) algorithm, specifically designed to learn and integrate category correlations for modeling semantic confusion in multi-label images, thereby facilitating more effective adaptive regularization. {We} employ a category-specific contrastive learning module to learn distinct feature representations for each category of a given image and their feature similarities. This approach enables us to model category correlations using instance-level feature similarities, allowing for the allocation of higher values to closely correlated categories and lower values to those less correlated, thus forming a soft label vector for adaptive regularization. Further enhancing our model, we learn prototype representations for each category and calculate the similarities between these prototypes and the features to establish more robust category correlations.  These correlations are then used to construct an additional soft label vector in an identical fashion. The integration of both soft label vectors, encompassing diverse and robust category correlations, allows them to be seamlessly incorporated into any existing multi-label recognition (MLR) methods \cite{chen2019multi,chen2019learning,lanchantin2021general} for effective calibration in a play-and-plug manner. As shown in Figure \ref{fig:mlcc-vis}, adding the DCLR algorithms to current models can lead to notably well-calibrated scores.

Currently, there exists no benchmark for MLCC evaluation, which severely prevents the development of this task. In this work, a unified evaluation benchmark is further built for fair MLCC comparisons and facilitates research in this field. Specifically, we first re-implement the traditional label smoothing \cite{muller2019does} and several leading confidence calibration algorithms \cite{liang2020improved, fernando2021dynamically, hebbalaguppe2022stitch, liu2022devil}, and adapt them to the MLCC task. For fair and comprehensive comparisons, we apply these algorithms and the proposed DCLR algorithms to three representative MLR models, i.e., SSGRL \cite{chen2019learning} and ML-GCN \cite{chen2019multi} that learns category-specific and holistic features for classification, and C-Tran~\cite{lanchantin2021general} that uses more advanced transformer networks. We use the traditional metrics like adaptive calibration error (ACE) \cite{nguyen2015posterior}, expected calibrated error (ECE) \cite{naeini2015obtaining}, maximum calibration error (MCE) \cite{hebbalaguppe2022stitch} and reliability diagram \cite{degroot1983comparison} for performance evaluation.

The contributions can be summarized into four folds. First, we extend confidence calibration from single-label recognition to the more practical and essential multi-label scenarios and present an in-depth analysis of the MLCC task. Second, we introduce a novel dynamic correlation learning and regularization algorithm that learns category correlations to model semantic confusion and integrates these correlations to regularize MLR model training to address the over-confidence dilemma in a play-and-plug manner. Third, we construct a fair and comprehensive MLCC evaluation benchmark, which can well evaluate the actual effects of each well-performing algorithm and facilitate the research in this field. Finally, we follow the unified benchmark to conduct extensive experiments to verify the effectiveness of the proposed algorithm. The unified benchmark, including the re-implemented codes and trained models of all re-implemented and our proposed DCLR algorithms, is available at \textbf{\url{https://github.com/wkml/MLCC-DCLR}}. 

\section{Related Works}
We review the related literature in terms of two main streams, i.e., multi-label image recognition and confidence calibration.   

\subsection{Multi-Label Image Recognition}
In recent years, multi-label image recognition (MLR) has garnered substantial academic interest, as evidenced by seminal works \cite{chen2022sst,chen2019learning,chen2024heterogeneous,nie2022multi,chen2022knowledge,lanchantin2021general,peng2023perception,pu2024dual}{\cite{ chen2020disentangling}}. This heightened attention to MLR underscores its pragmatic relevance and superiority over its single-label analog. Early MLR works propose to identify the local regions that contain more discriminative objects and their parts to learn more powerful feature representation. These works either pivot towards object proposals \cite{wei2016hcp,yang2016exploit} or resort to visual attention mechanisms \cite{ba2014multiple,chen2018recurrent,wang2017multi}. As a pioneer work, Wei et al., \cite{wei2016hcp} uses off-the-shelf algorithms to extract thousands of object proposes and aggregate their predicted scores to obtain the final MLR results. Despite achieving impressive progress, these algorithms merely consider visual features and do not consider label correlations. Indeed, there inherently exist strong label correlations in multi-label images, and their correlation can serve as additional knowledge to regularize MLR model training. Previous works, like \cite{wang2016cnn,wang2017multi} introduce recurrent neural networks (RNNs) or long short-term memory networks (LSTM) to learn contextualized feature representation that implicitly captures these correlations. Inspired by the graph propagation networks \cite{li2016gated,chen2021cross}, recent works \cite{chen2019multi,chen2019learning} postulate the use of structured graph representations to model label correlations in an explicit manner, which obviously facilitate MLR performance. For example, Chen et al. \cite{chen2019learning} first learn category-specific features, exploit label co-occurrence correlation to correlate these features and introduce graph neural networks to propagate information through the graph to learn contextualized features. 
Benefiting from the more powerful transformer networks \cite{vaswani2017attention,dosovitskiy2020image}{\cite{zhao2024two}}, more recent works further introduce transformer networks \cite{nguyen2021modular,lanchantin2021general} to learn more discriminative features or capture label correlations better and thus obtain better MLR performance.

\subsection{Confidence Calibration}
Confidence calibration of multi-class classification has been extensively studied for a long time. Confidence Calibration aims to make a classifier correctly quantify uncertainty or confidence associated with its instance-wise predictions. Many methods have been proposed~\cite{muller2019does,lin2017focal,mukhoti2020calibrating,liang2020improved,kumar2018trainable,hebbalaguppe2022stitch,liu2022devil,fernando2021dynamically}{\cite{shi2024label,an2024leveraging}} to correct the model confidence. Label smoothing (LS)~\cite{muller2019does} was proposed as a foundation technology of model calibration to prevent the network from becoming over-confident. Focal Loss (FL)~\cite{lin2017focal} focused training on a sparse set of hard examples and prevented many easy negatives from overwhelming the detector during training initially. In a follow-up study~\cite{mukhoti2020calibrating}, it was found that it can act as a model calibration method. Then, {Focal Loss with sample-dependent schedule (FLSD)}~\cite{mukhoti2020calibrating} was proposed as an improved focal loss to select the hyperparameter for model calibration automatically. {The difference between predicted confidence and accuracy (DCA)} ~\cite{liang2020improved}, a calibration method based on expected calibration error,  was proposed to correct confidence by adding the difference between predicted confidence and accuracy as an auxiliary loss. {Maximum Mean Calibration Error (MMCE)}~\cite{kumar2018trainable}, an RKHS kernel-based measure of calibration, is sound for perfect calibration that is minimized and whose finite sample estimates are consistent and enjoy fast convergence rates. Multi-class Difference in Confidence and Accuracy (MDCA)~\cite{hebbalaguppe2022stitch} acts as a novel auxiliary loss function to help a model achieve the same MDCA. {Margin-based Label Smoothing (MbLS)}~\cite{liu2022devil} provides a unifying constrained optimization perspective of current state-of-the-art calibration losses and then achieves a simple yet flexible generalization based on inequality constraints by imposing a controllable margin on logit distances. {Dynamically Weighted Balance Loss (DWBL)}~\cite{fernando2021dynamically} adjusted the model confidence with a class rebalancing strategy based on a class-balanced dynamically weighted loss. It can mitigate the class distribution imbalance issue in deep learning by assigning weights to different classes based on the class frequency and predicted probability of the ground-truth class. Adapting its weights automatically depending on the prediction scores allows a model to adjust for instances with varying difficulty levels, resulting in model calibration. 

Although the above methods have made significant progress in confidence calibration of the multi-class classification task, they still have not been explored under the setting of multi-label recognition. In this work, we extend them into the multi-label recognition task to validate their effectiveness and construct an evaluation benchmark by including them as baselines for a fair comparison. Besides, these methods do not consider category correlation that can well capture semantic confusion, leading to sub-optimal performance in the MLCC scenarios. To address this issue,  we propose a novel confidence calibration method, DCLR, to learn category correlations to model semantic confusion in multi-label images and integrate the correlations to build adaptive and more effective calibration.

\section{Dynamic Correlation Learning and Regularization}

\begin{figure*}[htbp]
  \centering
  \includegraphics[width=0.90\linewidth]{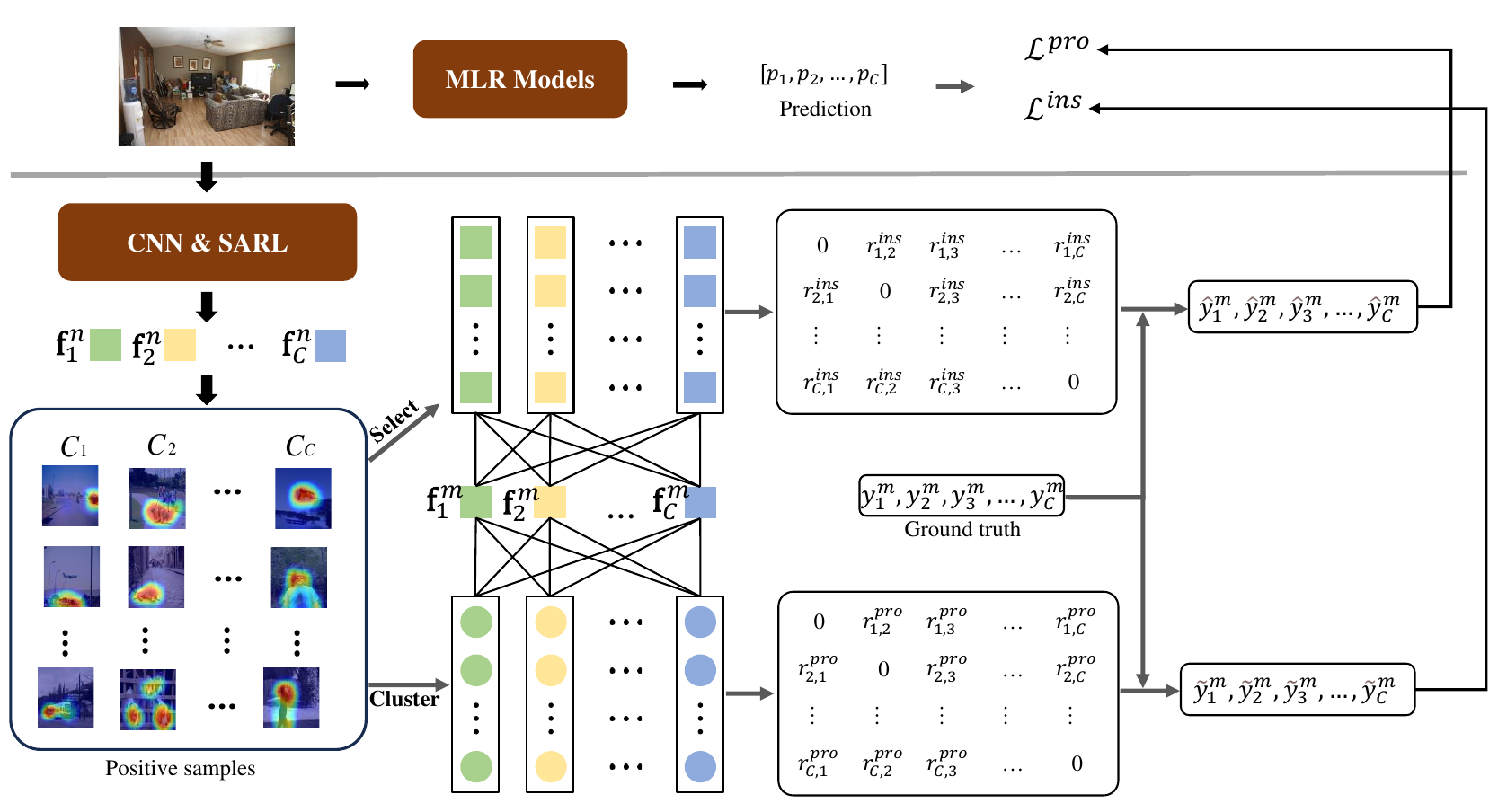}
  \vspace{-10pt}
  \caption{An overall illustration of the proposed DCLR algorithm and its integration into MLR models. Initially, an input image is processed through a backbone network followed by the SARL module to extract category-specific features. Subsequently, we compute instance-level and prototype-level correlation matrices by calculating the similarities between the extracted features and those of selected instance samples, as well as between the extracted features and prototype representations, to effectively model semantic confusion. Finally, we calculate the instance-level and prototype-level softened label vectors based on the respective correlation matrices and the ground truth labels. The softened label vectors are utilized for MLR model training.}
  \label{fig:framework}
\end{figure*}

MLR focuses on predicting labels of samples in a multi-class classification problem where each sample may belong to more than one class. Given a training set $\mathcal{D}=\{(\mathbf{x}^n, \mathbf{y}^n)\}^{N}_{n=1}$ and $C$ object categories, an MLR model aims to predict one or more object categories that truly belong to a given image. Here, $N$ is the total number of samples, $\mathbf{x}^n \in \mathbb{R}^{H \times W \times 3}$ denotes $n$-th image in $\mathcal{D}$, and $\mathbf{y}^{n}=\{y_{1}^{n}, \cdots, y_{C}^{n}\} \in \{1,0\}^{C}$ is the multi-hot encoded label vector. Most existing methods optimized a model to achieve this goal by applying the ground-truth multi-hot label vectors as supervisory signals with a binary cross-entropy loss function. The ground-truth multi-hot label vector comprises multiple hard labels of different categories where each hot is 0 or 1. Although these hard labels can define the categories to which an image belongs, optimizing a model with these hard labels often makes the model become over-confident, causing its output distribution to be untrustworthy and hard to associate with the accurate probability of correctness for the predictions~\cite{guo2017calibration, liu2022devil}. 

To mitigate the over-confident issue of neural networks, label smoothing (LS)~\cite{muller2019does, liu2022devil}, which is a form of output distribution regularization and a well-established single-label confidence calibration method, is often deployed to increase robustness and improve multi-class recognition problems. Label smoothing converts the hard labels into soft labels by applying a weighted average between the uniform distribution and the hard labels. It can be formulated as follows:
\begin{equation}
\label{equ:ls}
  \tilde{y}^{i}_{c} = \left\{
    \begin{aligned}
    1-\alpha, \quad & y^{i}_{c} = 1 \\
    \frac{\alpha}{C-1}, \quad & y^{i}_{c} = 0 
    \end{aligned}
  \right.
\end{equation}
It has been shown to calibrate the learned models implicitly by preventing the model from assigning the full probability mass to a single class and maintaining a reasonable distance between the logits of the ground-truth class and the other classes~\cite{pereyra2017regularizing,liu2022devil}. Intuitively, the label smoothing regularization can similarly be extended to multi-label recognition problems. It can be formulated as follows:
\begin{equation}
\label{equ:mls1}
  \tilde{y}^{i}_{c} = \left\{
    \begin{aligned}
    1-\alpha, \quad & y^{i}_{c} = 1 \\
    \frac{\alpha M}{C-M}, \quad & y^{i}_{c} = 0 
    \end{aligned}
  \right.
\end{equation}
where $M \leq C$ is the class number of the sample $\mathbf{x}^i$ and $\alpha \geq 0$ is a hyperparameter for output distribution regularization. If $M$ equals 1, Eq. (\ref{equ:mls1}) is equal to Eq. (\ref{equ:ls}). If $\alpha$ is set to 0, Eq. (\ref{equ:mls1}) equals the original labels. Besides, if $\alpha$ is too large, the model will fail to predict the ground-truth label.

The existing label smoothing algorithms distribute the penalty term, $\alpha$, uniformly across negative categories. However, these algorithms overlook the semantic confusion and correlation between similar object categories, a phenomenon naturally occurring in multi-label images. As a result, they fail to effectively regulate negative categories that bear a strong correlation with the positive categories. Thus, these highly-correlated categories remain confused with each other, leading to sub-optimal calibration performance in MLR scenarios. 

In this work, we posit a straightforward yet cogent assumption: negative categories with high similarity to positive ones should incur higher penalty values, while others should receive lesser penalties. To this end, we introduce the Dynamic Correlation Learning and Regularization (DCLR) algorithm, which is adept at learning category correlations to effectively model semantic confusion in multi-label images and incorporates these correlations to construct an adaptive and efficient regularization on MLR models. Figure \ref{fig:framework} illustrates the DCLR algorithm and its integration with existing MLR models. DCLR builds on a Category-Specific Contrastive Learning module to learn semantic-aware feature representation for each category. Then, the algorithm explores both instance-to-instance and instance-to-prototype similarities to model category correlations and computes softened label vectors based on these correlations. These vectors are subsequently employed to supervise existing MLR models. By implementing this adaptive regularization, DCLR effectively penalizes highly correlated categories, thereby more accurately resolving semantic confusion.

\subsection{Category-Specific Contrastive Learning}
To accurately gauge category correlation, it is essential to learn semantically-aware features for each category that satisfy two key criteria. Firstly, the feature vector for category $c$ should predominantly encapsulate the semantic information specific to that category, effectively minimizing the influence of information from other categories. Secondly, it is anticipated that features of correlated categories will exhibit a high degree of similarity, whereas those of uncorrelated categories will demonstrate significantly less similarity. To achieve this goal, we introduce a category-specific contrastive learning (CSFL) module. This module integrates category semantics to guide capturing the specific semantic information and is trained using a combination of classification and contrastive losses. It not only enhances focus on relevant semantic content but also guarantees the effective differentiation and alignment of features as per category correlations, ensuring that it adheres to the two aforementioned criteria.

Given an input image $I^m$, we first extract category-specific feature vectors via 
a standard backbone (e.g., ResNet101 \cite{he2016deep}) followed by semantic-embedded attention mechanism
as previous work \cite{chen2019learning}{, shorted as CNN\&SARL} . Formally, it can be represented as 
\begin{equation}
\{\mathbf f^m_1,\mathbf f^m_2,\cdots,\mathbf f^m_C\}=\phi(I^m).
\end{equation}
To ensure the learned features satisfy two key criteria, we introduce a combination of classification and contrastive losses. 

\noindent\textbf{Auxiliary classification loss. } To ensure that the feature vector $\mathbf{f}^m_c$ effectively encapsulates the semantic information pertinent to its category, we employ a classifier tasked with predicting the specific category $c$ solely based on the corresponding feature vector $\mathbf{f}^m_c$. This process is applied across all learned feature vectors, resulting in a probability distribution $\hat{\mathbf{p}}^m=\{\hat{p}^m_1, \hat{p}^m_2, \ldots, \hat{p}^m_C\}$. Subsequently, we leverage softened label vectors—specifically, $\tilde{\mathbf{y}}^m=\{\tilde{y}^m_1, \tilde{y}^m_2, \cdots, \tilde{y}^m_C\}$ and $\hat{\mathbf{y}}^m=\{\hat{y}^m_1, \hat{y}^m_2, \cdots, \hat{y}^m_C\}$ (the elaboration of these vectors can be found in Sections \ref{sec:ilcar} and \ref{sec:plcar}) to define the classification loss.
\begin{equation}\label{equ:acl}
\mathcal{L}_{acl} =\eta\sum_{m=1}^N(\ell(\hat{\mathbf{p}}^m, \hat{\mathbf{y}}^m) + \ell(\hat{\mathbf{p}}^m, \tilde{\mathbf{y}}^m)) 
\end{equation}
where 
\begin{equation}
\ell(\mathbf{p}, \mathbf{y})=\sum_{c=1}^{C}(y_c\log(p_c)+(1-y_c)\log(1-p_c))
\label{eq:cross-entropy}
\end{equation}
In this context, we set $\eta$ to 0.5 to align the magnitude of {the traditional cross-entropy loss which is a common-used classification loss}. 

\noindent\textbf{Category-specific contrastive loss.} Another criterion we consider is the expectation that features within correlated categories will show a high degree of similarity while those in uncorrelated categories will display less similarity. However, due to the lack of precise annotations detailing these category similarities, accurate supervision signals to learn these correlations are unavailable. In this work, we introduce a category-specific contrastive loss to leverage existing label annotations for learning these correlations. For any two images and a given category, our approach seeks to draw the corresponding features closer when the given category is present in both images and to distance them otherwise. Since objects from the same category in different images typically share similar visual characteristics \cite{chen2022structured}, it can effectively learn visual correlations across different categories. 

Formally, given two images $m$, $n$, and category $c$, we compute their similarity via the cosine distance
\begin{equation}
s_{c}^{m,n}=\mathrm{cosine}(\mathbf f_{c}^m,\mathbf f_{c}^n)=\dfrac{\mathbf f_{c}^m\cdot\mathbf f_{c}^n}{||\mathbf f_{c}^m||\cdot||\mathbf f_{c}^n||}.
\end{equation}
It is expected that $s_{c}^{m,n}$ to be high if both images $m$ and $n$ contain category $c$ and to be low otherwise. To this end, we formulate the category-specific contrastive loss as 
\begin{equation}
\label{cll}
\mathcal{L}_{c}=\sum_{m=1}^N\sum_{c=1}^C\sum_n\ell_c^{m,n},
\end{equation}
where
\begin{equation}
\ell^{m,n}_c=\begin{cases}1-s_{c}^{m,n}&y^i_c=1,y^j_c=1\\ 1+s_{c}^{m,n}&otherwise.\end{cases}
\end{equation}

{
Therefore, the final loss of the category-specific contrastive learning (CSCL) module is defined as follows:
}
\begin{equation}
\label{equ:final}
\mathcal{L}_{cscl} =\mathcal{L}_{acl}+\mathcal{L}_{c}
\end{equation}

\noindent\textbf{{Traning Algorithm.}}
{The overall training algorithm of our CSCL is presented in Algorithm~\ref{algorithm:1}. As introduced before, at each training step, Lines 4-6 take charge of producing soft labels at the instance level and prototype level. Then, Lines 7-9 calculate the auxiliary classification loss $\mathcal{L}_{acl}$, category-specific contrastive loss $\mathcal{L}_{c}$, and the final loss $\mathcal{L}_{cscl}$. The training procedure will continue over the data epochs till convergence. After the CSCL is trained, it can be used to generate soft labels to guide the training of various MLR models.}

\begin{algorithm}[t]
    \caption{Training Procedure of Category-Specific Contrastive Learning (CSCL)}
    \hspace*{0.02in} 
    {\bf Input:} Training image dataset $\mathcal{D}$; 
     
    \hspace*{0.02in} 
    {\bf Output:} well-trained CNN\&SARL module $\phi$; 
    
    \begin{algorithmic}[1]
    \STATE initialize CNN\&SARL module $\phi$;
    \REPEAT 
        \FORALL {$I^m \in \mathcal{D}$ }
            \STATE $\{\mathbf f^m_1,\mathbf f^m_2,\cdots,\mathbf f^m_C\}=\phi(I^m)$
            \STATE obtain instance-level soft label vector $\hat{\mathbf{y}}^m$ according to Equation (\ref{equ:isl})
            \STATE obtain prototype-level soft label vector $\tilde{\mathbf{y}}^m$ according to Equation (\ref{equ:psl})
            \STATE compute $\mathcal{L}_{acl}$ according to Equation (\ref{equ:acl})
            \STATE compute $\mathcal{L}_{c}$ according to Equation (\ref{cll})
            \STATE $\mathcal{L}_{cscl} =\mathcal{L}_{acl}+\mathcal{L}_{c}$
            \STATE update the model parameters of $\phi$ by minimizing $\mathcal{L}_{cscl}$
        \ENDFOR
    \UNTIL {$\phi$ convergence}
   
    \RETURN the CNN\&SARL module $\phi$
    \end{algorithmic}
    \label{algorithm:1}
\end{algorithm}

\subsection{Instance-Level Correlation-Aware Regularization}
\label{sec:ilcar}
After acquiring the category-specific feature vector, we can calculate correlations between various categories through category-specific feature similarities, resulting in the formation of $\mathbf{R}^{ins}_{}$. In this matrix, the element $r^{ins}_{ij}$ positioned in the $i$-th row and $j$-th column signifies the correlation between {categories $i$ and $j$}. 
Subsequently, the generated matrix facilitates the derivation of a soft label vector, which is instrumental in training the MLR models.

Given an image $m$ and category $c$, to compute the correlation of $c$ with $c'$, we first retrieve a subset of images that have positive label $c'$. The correlation value can be computed by    
\begin{equation}
r^{ins}_{cc'} = \sum_{t=1}^T\mathrm{cosine}(\mathbf f_{c}^m,\mathbf f_{c'}^{n_t})
\end{equation}
Here, $T$ is the number of retrieved images. For each image $m$, we compute the correlation values across all {categories} $c$ and $c'$ to obtain the correlation matrix $\mathbf{R}^{ins}_{m}$.

To eliminate the influence of the label itself in the penalty term, we need to mask the diagonal of the correlation matrix, which can be implemented by 
\begin{equation}
r^{ins}_{cc'}=
\begin{cases}
 -\infty, &c=c' \\
r^{ins}_{cc'}, &otherwise.
\end{cases}
\end{equation}
Subsequently, we normalize the similarity scores among all categories using a $\mathrm{softmax}$ function as follows:
\begin{equation}
r^{ins}_{cc'} = \dfrac{\exp(r^{ins}_{cc'})}{\sum_{k} \exp(r^{ins}_{ck})}.
\end{equation}
Finally, we can obtain {a} soft label vector according to the instance-level correlation matrices $\mathbf{R}^{m}_{ins}$ as follows:
\begin{equation}
\label{equ:isl}
  \hat{y}^{m}_{c} = \left\{
    \begin{aligned}
    1-\alpha, \quad & y^{m}_{c} = 1 \\
    \sum_{k=1}^{C} \alpha r^{ins}_{kc}y^{m}_{k}, \quad & y^{m}_{c} = 0 
    \end{aligned}
  \right.
\end{equation}
We repeat the process for all categories and obtain instance-level soft label vector $\hat{\mathbf{y}}^m=\{\hat{y}^m_1, \hat{y}^m_2, \cdots, \hat{y}^m_C\}$.

\subsection{Prototype-Level Correlation-Aware Regularization}
\label{sec:plcar}
Although the above instance-level correlation-aware regularization can be deployed as an improved label smoothing, it only captures relatively local inter-class correlations because only a subset of images is sampled to compute the category-specific correlations for each input image. To incorporate global inter-class correlations into our framework to build more robust smoothed labels for the model calibration, we further model the category-specific correlations globally by introducing category prototypes that can represent the global features of categories.

To achieve prototype-level correlation-aware regularization, for each category $c$, we first extract all instance-level features from all the images that have been labeled the category $c$ as one of their labels. Then, we apply $K$-means algorithm~\cite{hartigan1979algorithm} to cluster instance-level features into $K$ prototype-level features $\{\mathbf p^1_{c},\mathbf p^2_{c},...,\mathbf p^K_{c}\}$. Here, the $K$ is set to 10 since the images belonging to any category $c$ have visual variations. For example, objects belonging to the same category but from different images may have different appearances. Therefore, we cluster multiple prototype-level features to account for visual variations for any category $c$.

Given an image $m$ and category $c$, we can compute the correlation value in an identical manner, formulated as 
\begin{equation}
\label{sproto}
r^{pro}_{cc'}=\sum_{k=1}^K\mathrm{cosine}(\mathbf{f}_{c}^{m},\mathbf{p}_{c}^{k}).
\end{equation}
Similarly, we compute the correlation values for all categories $c$ and $c'$ according to equation \ref{sproto}, followed by assigning {values} of the diagonal as $-\infty$ and normalizing them with the $\mathrm{softmax}$ function to obtain the prototype-level correlation matrix $\mathbf{R}^pro_{m}$. Afterwards, we apply it to obtain the soft label vector as
\begin{equation}
\label{equ:psl}
  \tilde{y}^{m}_{c} = \left\{
    \begin{aligned}
    1-\alpha, \quad & y^{m}_{c} = 1 \\
    \sum_{k=1}^{C} \alpha r^{pro}_{kc}y^{m}_{k}, \quad & y^{m}_{c} = 0 
    \end{aligned}
  \right.
\end{equation}
Finally, we can obtain prototype-level soft label vector $\tilde{\mathbf{y}}^m=\{\tilde{y}^m_1, \tilde{y}^m_2, \cdots, \tilde{y}^m_C\}$.

\subsection{Regularization on Existing MLR Models}
When the DCLR algorithm is trained, it can be applied to generate softened label vectors instead of traditional one-hot label vectors. These softened vectors are then used to supervise the training of MLR models. Given an input image $n$, it is fed into the MLR model to predict the probability distribution $\mathbf{p}^n=\{p^n_1, p^n_2, \ldots, p^n_C\}$ and fed into the DCLR model to obtain the soften label vectors, i.e., $\mathbf{p}_{ins}^n=\{y^n_{ins, 1}, y^n_{ins, 2}, \ldots, y^n_{ins, C}\}$ and $\mathbf{p}_{pro}^n=\{y^n_{pro, 1}, y^n_{pro, 2}, \ldots, y^n_{pro, C}\}$. The classification loss function can be defined as 
\begin{equation}
\mathcal{L}_{cls} =\eta(\mathcal{L}^{ins}+\mathcal{L}^{pro}) 
\label{eq:cls-loss}
\end{equation}
where
\begin{equation}
\label{equ:loss}
  \begin{aligned}
    \mathcal{L}^{ins}&=\sum_{m=1}^N(\ell(\mathbf{p}^m, \hat{\mathbf{y}}_m) \\
    \mathcal{L}^{pro}&=\sum_{m=1}^N(\ell(\mathbf{p}^m, \tilde{\mathbf{y}}_m)
    \end{aligned}
\end{equation}
In this formulation, $\ell(\mathbf{p}, \mathbf{y})$ is defined according to Equation \ref{eq:cross-entropy}, with $\eta$ consistently set to 0.5.

To address the issue of overconfidence in existing MLR models \cite{chen2019learning,chen2019multi,lanchantin2021general}, we substitute the traditional cross-entropy loss, which is based on one-hot label vectors, with the classification loss as defined in Equation \ref{eq:cls-loss}. Apart from replacing the classification loss, all other aspects, such as network details and training processes, remain identical to those in the original works.

\section{Unified Evaluation Benchmark}
In this section, we present the selected multi-label image recognition methods, competing algorithms, and datasets involved in constructing the unified multi-label confidence calibration (MLCC) evaluation benchmark. Then, we introduce the unified evaluation protocols of the benchmark for fair comparison.

\subsection{Selected Multi-Label Image Recognition Methods}
Most calibration algorithms are implemented on different baselines or targeted in different scenarios, preventing fair evaluation. To evaluate calibration algorithms fairly on the multi-label image recognition task, we choose three multi-label recognition methods as the backbones, including SSGRL \cite{chen2019learning}, ML-GCN \cite{chen2019multi}, and C-Tran~\cite{lanchantin2021general}. A brief introduction of these methods is as follows:
\begin{itemize}
\item SSGRL \cite{chen2019learning}: It is a semantic-specific graph representation learning framework with two key modules. The first key module is a semantic decoupling module that guides semantic-specific representation learning by incorporating category semantics. Another module is a semantic interaction module that correlates these semantic-specific representations with a graph built on the statistical label co-occurrence and explores their interactions via a graph propagation mechanism.
\item ML-GCN \cite{chen2019multi}: It is a multi-label classification model based on Graph
Convolutional Network (GCN) to capture and explore label dependencies where objects normally co-occur in an image. It builds a directed graph over the object labels and applies learnable GCN to map the label graph into a set of inter-dependent object classifiers that are applied to the image descriptors extracted by another sub-net. Besides, it also applies a new re-weighted scheme to construct an effective label correlation matrix to guide the optimization of the learnable GCN. 
\item C-Tran~\cite{lanchantin2021general}: C-Tran is a general multi-label image classification framework to leverage transformers~\cite{vaswani2017attention} to exploit the complex dependencies among visual features and labels. It consists of a Transformer encoder trained to predict a set of target labels given an input set of masked labels and visual features from a CNN. Besides, a novel label masking training objective is also proposed to use a ternary encoding scheme to represent the label states as positive, negative, or unknown during training.
\end{itemize}
The reason that we chose these three multi-label image recognition methods as the backbones is their different representative characteristics for different mainstream multi-label image recognition methods. SSGRL is a multi-label image recognition method with category-specific features, while ML-GCN models holistic features for classification. C-Tran~\cite{lanchantin2021general} is a multi-label image recognition model driven by the transformer which is a new foundation model.

\subsection{Competing Algorithms}
For fair evaluation, we choose some classical algorithms and reimplement them on our selected backbones. A simple overview of different competing algorithms is introduced as follows:

\begin{itemize}
\item NLL. The negative log-likelihood loss (NLL) is useful for training a classification problem with multiple classes. Since multi-label visual recognition is a special multiclass classification task, it is adopted for one of our baselines.
\item LS\cite{muller2019does}. Label smoothing (LS) is a foundation technology of model calibration since it can prevent the network from becoming over-confident. Besides, our work is also to improve upon it.
\item FL\cite{lin2017focal}. Focal Loss (FL) focuses training on a sparse set of hard examples and prevents many easy negatives from overwhelming the detector during training initially. In a follow-up study~\cite{mukhoti2020calibrating}, researchers found it also can be applied to model calibration. Therefore, we treat it as one of the baselines for comparison.
\item FLSD\cite{mukhoti2020calibrating}. {Focal Loss with sample-dependent schedule (FLSD)} is an improved focal loss by automatically selecting the hyperparameter for model calibration.
\item DCA\cite{liang2020improved}. {The difference between predicted confidence and accuracy (DCA)} is a model calibration method based on expected calibration error by adding the difference between predicted confidence and accuracy as an auxiliary loss.
\item MMCE\cite{kumar2018trainable}. {Maximum Mean Calibration Error (MMCE)} is an RKHS kernel-based measure of calibration. It is efficiently trainable with the negative likelihood loss without elaborate hyper-parameter tuning. MMCE is also sound for perfect calibration that is minimized and whose finite sample estimates are consistent and enjoy fast convergence rates.
\item MDCA\cite{hebbalaguppe2022stitch}. Multi-class Difference in Confidence and Accuracy ( MDCA ) is a novel auxiliary loss function to help a model achieve the same MDCA. 
\item MbLS\cite{liu2022devil}. The authors in {Margin-based
Label Smoothing (MbLS)}~\cite{liu2022devil} first provide a unifying constrained optimization perspective of current state-of-the-art calibration losses and then propose a simple yet flexible generalization based on inequality constraints by imposing a controllable margin on logit distances.  
\item DWBL\cite{fernando2021dynamically}. {Dynamically Weighted Balance
Loss (DWBL)} is a class rebalancing strategy based on a class-balanced dynamically weighted loss to mitigate the class distribution imbalance issue in deep learning. It assigns weights to different classes based on the class frequency and predicted probability of the ground-truth class. Adapting its weights automatically depending on the prediction scores allows a model to adjust for instances with varying difficulty levels resulting in model calibration.  
\end{itemize}

\subsection{Datasets}

We use two famous and publicly available datasets, MS-COCO \cite{lin2014microsoft} and Visual Genome~\cite{krishna2017visual} as the benchmark datasets as done in many previous multi-label recognition works\cite{chen2019learning, zhu2017learning} so that we can compare various algorithms fairly and consistently. Although Pascal VOC 2007 \cite{everingham2010pascal} is also widely used as one of the benchmarks in various multi-label recognition works, we do not use Pascal VOC 2007 \cite{everingham2010pascal} as one of the benchmark datasets since it only has 20 common categories and may not exist the over-confident issue.   

The MS-COCO \cite{lin2014microsoft} dataset was initially designed for object detection and segmentation. Recently, it also has been adopted directly for evaluating multi-label image recognition due to its fine-grained object annotation. This dataset comprises 122,218 images and contains 80 common categories. These images are divided into a training set consisting of 82,081 images and a validation set comprising 40,137 images. It must be noticed that the ground truth annotations for the test set are currently unavailable. Therefore, all models reported in this paper are trained on the training set and evaluated on the validation set.


Visual Genome~\cite{krishna2017visual} contains 108,249 images with densely annotated objects, attributes, and relationships. These images are from the intersection of the YFCC100M \cite{thomee2016yfcc100m} and MS-COCO \cite{lin2014microsoft} and cover 80,138 categories. Each image has an average of 35 objects in the Visual Genome. Therefore, the Visual Genome is a good test bed for measuring the performance of the multi-label recognition task and the effect of model calibration. Since most categories have very few samples, we merely consider the 500 most frequent categories and we randomly select 10,000 images as the test set and the rest 98,249 images as the training set.

\subsection{Evaluation Metrics} 
Our framework for multi-label confidence calibration is focused on achieving both accuracy and calibration. To evaluate the accuracy, we follow previous works\cite{wei2015hcp, yang2016exploit} to adopt the mean average precision (mAP) across all categories. In terms of calibration, we adopt several widely-used metrics, including the expected calibration error (ECE)\cite{naeini2015obtaining}, adaptive ECE (ACE)\cite{nixon2019measuring}, and maximum calibration error (MCE)\cite{hebbalaguppe2022stitch}. The ECE measures the difference in expected accuracy and expected confidence. The ACE measures the difference between accuracy and confidence in an adaptive scheme that spaces the bin intervals so that each contains an equal number of predictions. The MCE measures the maximum difference between average confidence and accuracy.

\begin{table*}[t]
  \centering
  \caption{The performance of our DCLR and various baselines on three different MLC models and two common-used benchmarks. The best and second best results are \textbf{highlighted} and \underline{underlined}.}
  \begin{tabular}{c|c|cccc|cccc|cccc}
  \hline
  \centering \multirow{2}{*}{Datasets} & \multirow{2}{*}{Method} & \multicolumn{4}{c|}{SSGRL~\cite{chen2019learning}} & \multicolumn{4}{c|}{ML-GCN~\cite{chen2019multi}}  & \multicolumn{4}{c}{C-Tran~\cite{lanchantin2021general}} \\ 
  \cline{3-14}
  \centering & & mAP & ACE & ECE & MCE & mAP & ACE & ECE & MCE & mAP & ACE & ECE & MCE \\
  \hline
  \hline
  \multirow{10}{*}{MS-COCO}
        & NLL & 84.0 & 3.692 & 4.108 & 154.069 &83.0  &4.143  &5.072  &232.622  &83.5  &2.414  &3.675  &154.668\\
        & LS\cite{muller2019does} & 83.9 & 2.104 & \underline{1.741} & \underline{69.516} &83.5  &\underline{1.436}  &\underline{2.165}  &55.906  &83.8  &\underline{1.146}  &\underline{1.866}  &\underline{55.621}\\
        & FL\cite{lin2017focal} & 84.0 & 3.049 & 4.734 & 81.670 &83.0  &4.011  &5.007  &65.005  &83.6  &2.012  &3.023  &130.523\\
        & FLSD\cite{mukhoti2020calibrating} & 84.1 & 3.546 & 3.858 & 133.999 &82.9  &4.196  &4.967  &\underline{48.713}  &83.7  &1.915  &2.775  &100.296\\
        & DCA\cite{liang2020improved} & 84.1 & 3.376 & 3.636 & 127.368 &82.8  &3.931  &4.131  &120.012  &83.7	&2.180	&3.891	&130.962\\
        & MMCE\cite{kumar2018trainable} & 84.1 & \underline{1.575} & 2.951 & 130.991 &82.8  &3.523  &3.840  &121.866  &83.7 &2.111 &2.874	&109.911\\
        & MDCA\cite{hebbalaguppe2022stitch} & 84.1 & 3.449 & 3.834 & 148.859 &82.9  &3.880  &4.067  &125.018  &83.7	&1.942	&3.668	&125.107\\
        &MbLS\cite{liu2022devil} & 83.3 & 2.304 & 2.188 & 96.764 &83.3  &2.923  &2.954  &67.812  &83.9	&1.808	&2.582	&108.604\\
        & DWBL\cite{fernando2021dynamically} & 84.0 & 3.181 & 4.345 & 79.217 &81.3  &9.726  &11.170  &115.477  &83.6  &2.195  &2.519  &102.985\\
        & Ours & 84.0 & \textbf{1.459} & \textbf{1.503} & \textbf{59.018} &83.5  &\textbf{1.037}  &\textbf{1.427}  &\textbf{26.411}  &83.8	&\textbf{0.862}	&\textbf{1.489}	&\textbf{51.854}\\ 
  \hline
  \hline
  \multirow{10}{*}{Visual Genome}
        & NLL &51.4 &5.053 &5.199	&52.272 &48.6 &4.804 &4.811 &200.376  &51.6	&4.338 &4.678 &44.497 \\
        & LS\cite{muller2019does} &51.4 &\underline{4.400} &\underline{3.721} &\underline{35.902}  &48.7 &\underline{3.643} &\underline{3.892} &\underline{108.051} &51.6	&2.985 &\underline{2.673} &\underline{33.170} \\
        & FL\cite{lin2017focal} &51.4	&4.814	&4.914	&42.745  &48.7  &3.794  &3.907  &187.978  &51.8 &4.012 &3.918 &39.124 \\
        & FLSD\cite{mukhoti2020calibrating} &51.5	&4.541	&4.874	&40.798  &48.7  &3.587  &3.876  &189.701  &51.8 &2.981 &3.513 &40.325 \\
        & DCA\cite{liang2020improved} &51.6 &4.565 &4.820 &51.406 &48.7  &4.518  &4.866  &178.643  &51.6 &3.824 &4.198 &39.908 \\
        & MMCE\cite{kumar2018trainable} &51.5	&4.560 &4.851 &46.015  &48.6  &4.015  &4.598  &158.092  &51.7 &4.032 &4.070 &44.307  \\
        & MDCA\cite{hebbalaguppe2022stitch} &51.4	&4.786 &5.054 &52.529  &48.6  &4.076  &4.879  &150.159  &51.7 &4.205 &4.273 &45.286  \\
        & MbLS\cite{liu2022devil} &51.6 &4.741 &3.907 &45.676 &48.7  &3.987  &4.675  &149.802  &51.7 &\underline{2.964} &4.329 &57.031 \\
        & DWBL\cite{fernando2021dynamically} &51.5 &4.494 &3.974 &37.851 &48.6  &4.498  &4.523  &149.800  &51.6  &3.415  &3.298  &36.267  \\
        & Ours &51.4 &\textbf{3.548} &\textbf{2.887} &\textbf{33.248}  &48.8 &\textbf{3.580} &\textbf{3.752} &\textbf{72.487}  &51.6  &\textbf{2.608}  &\textbf{2.480}  &\textbf{28.286} \\
  \hline
  \end{tabular}
  \label{tab:all}
\end{table*}

\section{Experiments}
In this section, we present the implementation details and then report and analyze the experimental results of our method and various baselines. Besides, we also conduct ablation studies to validate the effects of different components in our algorithm. Finally, we show the universality of our DCLR on downstream applications.

\subsection{Implementation details}
We use ResNet-101~\cite{he2016deep}, which is a convolutional neural network that has 101 layers as the backbone network for feature extraction of DCLR. For better generalization, we first use pre-trained weights to initialize the backbone network and then fine-tune it with the data of the downstream task. We train the DCLR model with the Adam~\cite{kingma2014adam} optimizer for 20 epochs. The initial learning rate, beta1, and beta2 are set to $10^{-5}$, 0.999 and 0.9, respectively. We reduce the learning rate by 10 times at the 10th epoch. To train the MLR models, we follow the training settings of the original SSGRL \cite{chen2019learning}, ML-GCN \cite{chen2019multi}, and C-Tran \cite{lanchantin2021general} works.



\begin{figure*}[htbp]
  \centering
  \includegraphics[width=1.0\linewidth]{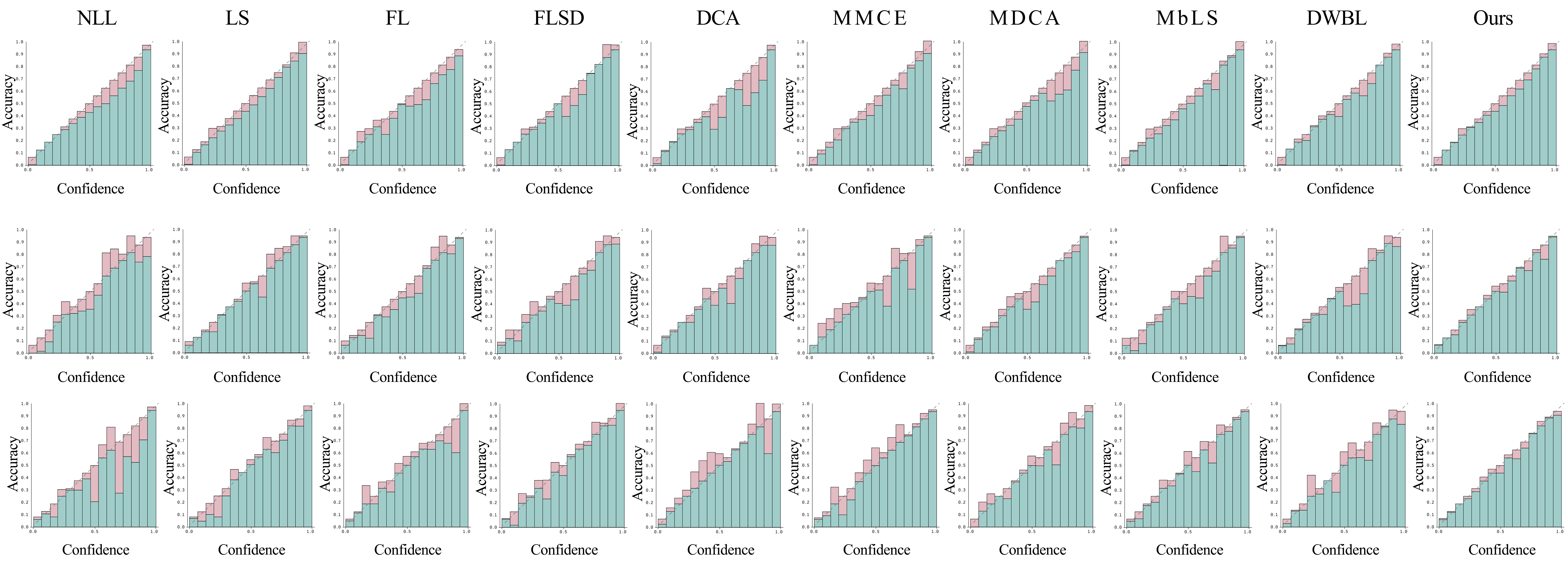}
  \vspace{-20pt}
  \caption{The reliability diagrams for SSGRL, ML-GCN, and C-Tran models without and with the existing competing and proposed DCLR algorithms on the MS-COCO dataset. The results for these models are systematically organized: SSGRL is shown in the first row, ML-GCN in the second row, and C-Tran in the last row.}
  \label{fig:method_wise}
\end{figure*}

\subsection{Comparisons of DCLR with Various Baselines}
In this part, we comprehensively evaluate the performance of both re-implemented and our proposed DCLR algorithms, under the fair evaluation benchmark for a thorough and fair comparison. As presented in Table~\ref{tab:all}, our proposed DCLR algorithm surpasses all competing leading calibration algorithms in terms of ACE, ECE, and MCE metrics across all three models \cite{chen2019learning,chen2019multi,lanchantin2021general}. This superior improvement is consistent on both the MS-COCO \cite{lin2014microsoft} and Visual Genome~\cite{krishna2017visual} datasets. Notably, these advancements are achieved without any detriment to the mAP metric, which is crucial for measuring classification performance.

The results are presented in Table \ref{tab:all}. In our detailed analysis of the MS-COCO dataset, we observed that existing confidence calibration algorithms significantly enhance the ACE, ECE, and MCE metrics in most scenarios. Our newly proposed DCLR algorithm outperforms these existing solutions across all metrics, demonstrating notable improvements. Specifically, it achieves ACE, ECE, and MCE scores of 1.459, 1.503, and 59.018 for the SSGRL model; 1.037, 1.427, and 26.411 for the ML-GCN model; and 0.862, 1.489, and 51.854 for the C-Tran model. Compared with the second best-performing LS algorithm, these results translate to relative reductions in ACE, ECE, and MCE of 30.7\%, 13.7\%, and 15.1\% for the SSGRL model; 27.8\%, 34.1\%, and 52.8\% for the ML-GCN model; and 24.8\%, 20.2\%, and 6.8\% for the C-Tran model, respectively. These results strongly underscore the superior performance of the DCLR algorithm. 

In the case of the Visual Genome dataset, similar phenomena emerge. Adding {the} DCLR algorithm leads to a notable decrease in the ACE, ECE, and MCE metrics. For the SSGRL model, the metrics improved from 4.400, 3.721, and 35.902 to 3.548, 2.887, and 33.248. For the ML-GCN model, the figures changed from 3.643, 3.892, and 108.051 to 3.580, 3.752, and 72.487. Additionally, for the C-Tran model, the metrics were reduced from 2.964, 2.673, and 33.170 to 2.608, 2.480, and 28.286. These results further confirm the effectiveness of the DCLR algorithm in enhancing calibration performance.

For more comprehensive evaluations,  we also provide reliability diagrams that graphically illustrate the discrepancy between statistical predictive confidence and actual accuracy. The closer this curve aligns with the diagonal line, the better the calibration performance. Figure \ref{fig:method_wise} showcases these diagrams for SSGRL, ML-GCN, and C-Tran models, both with and without the application of existing and our proposed DCLR calibration algorithms. The models without calibration significantly deviate below the diagonal, suggesting a severe overconfidence dilemma. Utilizing existing calibration methods moderately improves model alignment. In contrast, the models calibrated with our DCLR algorithm demonstrate near-perfect alignment with the diagonal, implying a near-ideal state of calibration.

It is essential for confidence calibration algorithms not to compromise the classification performance, a key aspect of multi-label recognition (MLR) tasks. Thus, we further present the mAP metric, the most commonly used measure in MLR evaluation. As shown in Table \ref{tab:all}, the proposed DCLR and most existing algorithms yield competitive mAP scores comparable to those achieved without calibration across various models and datasets. Notably, we find traditional LS outperforms most existing algorithms. This disparity in performance may be attributed to the fact that most existing algorithms are primarily designed for single-label scenarios and thus exhibit poor performance for the multi-label counterparts. In contrast, the straightforward approach of label smoothing regularization is more adaptable across various settings, resulting in enhanced performance.

\subsection{Ablation Study}



To deeply analyze the DCLR algorithm and the actual contributions of each crucial module, we further present extensive ablative experiments. Here, all experiments are conducted using the SSGRL baseline on the MS-COCO dataset. 

\begin{figure}[!t]
  \centering
  \includegraphics[width=1.0\linewidth]{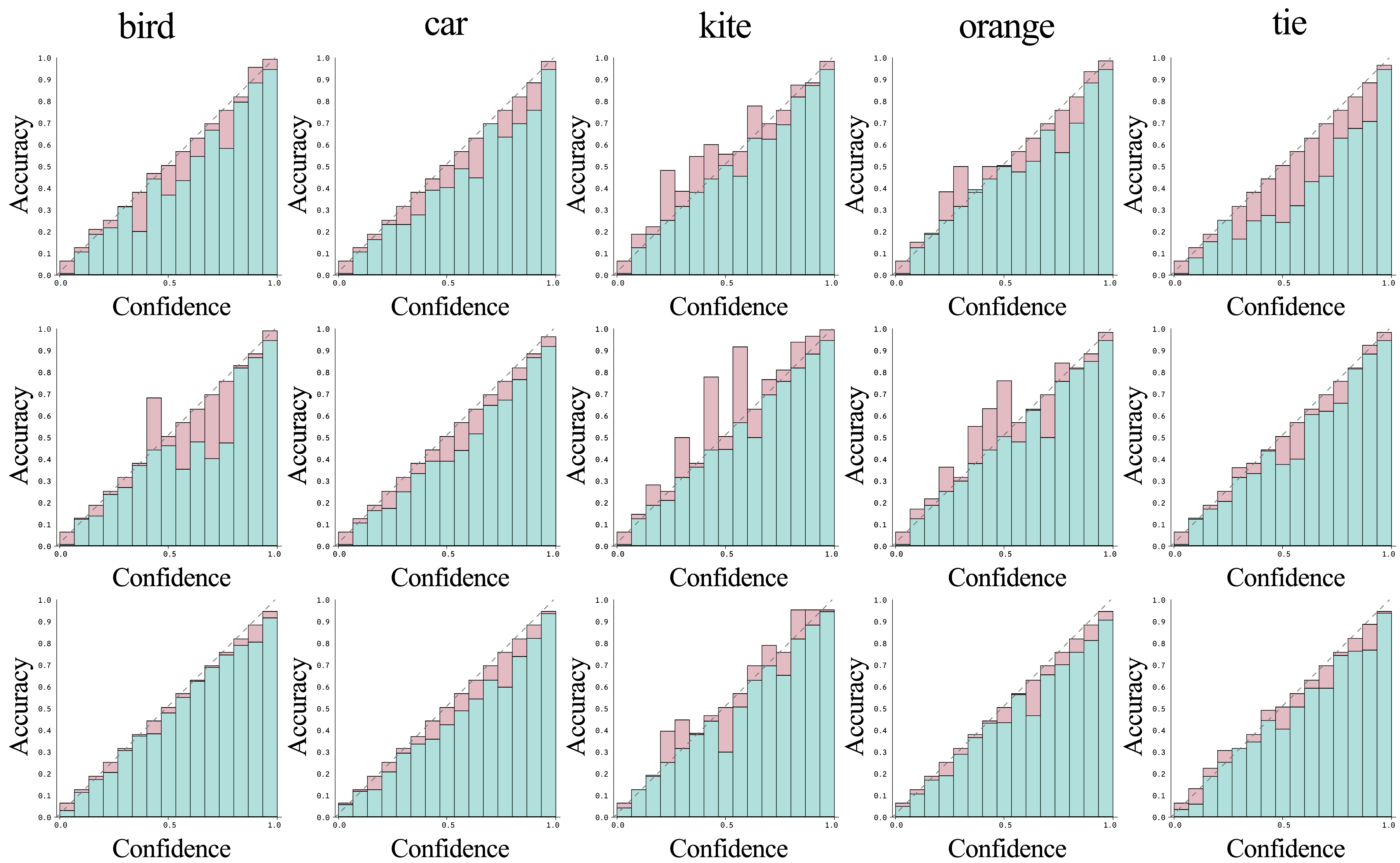}
  \vspace{-20pt}
  \caption{The reliability diagrams of each category using no calibration algorithm, as well as using the LS and proposed DCLR algorithms. The results for these algorithms are organized sequentially: the results using no calibration algorithm are in the first row, those using LS in the second row, and those with the DCLR algorithm are displayed in the last row. The evaluations are conducted using the SSGRL model on the MS-COCO dataset.}
  \label{fig:class_wise}
\end{figure}

\subsubsection{Analysis of DCLR}

The key technical contribution of this work is the DCLR algorithm, which incorporates category correlations to effectively resolve semantic ambiguities, thereby enhancing regularization. We highlight its efficacy by comparing it with results from uncalibrated models (NLL) and those calibrated with the LS algorithm. As presented in Table \ref{tab:all}, utilizing the DCLR algorithm notably enhances performance compared with both the NLL and LS-calibrated models. This improvement is further mirrored in the reliability diagrams shown in Figure \ref{fig:method_wise}. For a more thorough comparison, we also present category-specific reliability diagrams in Figure \ref{fig:class_wise}, where we have chosen five categories from MS-COCO for a concise illustration. The DCLR algorithm significantly narrows the gap between statistical predictive confidence and actual accuracy, indicating more accurately calibrated scores.

These evaluations demonstrate the effectiveness of the proposed DCLR algorithm as a whole. Actually, it contains both instance-level and prototype-level correlation-aware regularization. In the following, we further present an in-depth analysis of these two modules.

\subsubsection{{Effects of CSCL, ILCAR, and PLCAR}}
{To investigate the effects of category-specific contrastive learning (CSCL), instance-level correlation-aware regularization (ILCAR), and prototype-level correlation-aware regularization (PLCAR), we conduct an ablation study on the SSGRL model and the MS-COCO dataset by enabling/disabling them. The experimental results are shown in Table~\ref{tab:cscl}. We can observe that all three different components can help the model achieve better calibration. To be specific, Comparing Ours DCLR with Ours DCLR w/o $\mathcal{L}_c$, we can see that CSCL reduces the ACE, ECE, and MCE by 0.439, 0.144, and 9.57, respectively. Similarly, PLCRA reduces the ACE, ECE, and MCE by 0.33, 0.117, and 5.01 compared Ours DCLR with Ours DCLR w/o $\mathcal{L}^{pro}$ while ILCRA reduces the ACE, ECE, and MCE by 0.342, 0.09, and 8.275 compared Ours DCLR with Ours DCLR w/o $\mathcal{L}^{ins}$, respectively. These results show the effectiveness of these three components in improving model confidence. Moreover, when these three components are applied simultaneously, the model achieves the best calibration performance, showing the effectiveness of our DCLR.
} 

\begin{table}[htbp]
\caption{The effects of CSCL, ILCAR, and PLCAR on the SSGRL model.}
\centering
\begin{tabular}{c|cccc}
\hline
&  mAP & ACE & ECE & MCE \\ \hline
Ours DCLR w/o $\mathcal{L}_c$ &84.1 &1.897 &1.647 &68.588  \\ 
Ours DCLR w/o $\mathcal{L}^{ins}$ &84.0 & 1.801 & 1.593 & 67.293 \\  
Ours DCLR w/o $\mathcal{L}^{pro}$ &83.9 & 1.789  & 1.620 & 64.028  \\
Ours DCLR & 84.0 & 1.459 & 1.503 & 59.018 \\  
\hline
\end{tabular}
\label{tab:cscl}
\end{table}

\subsubsection{{Analysis of the hyper-parameter $\alpha$}}
{The hyper-parameter $\alpha$ is used for output distribution regularization. To measure its influence on the performance and set up proper value, we conduct an ablation study on MS-COCO with the backbone SSGRL by setting it as 0.02, 0.03, 0.05, 0.1. The experiment results are shown in Table~\ref{tab:alpha}. From the results, we can observe that all different $\alpha$ settings can achieve competitive performance on various metrics while the best and state-of-the-art performance is achieved by setting $\alpha$ as 0.05. Therefore, we choose 0.05 as the default value of the hyper-parameter $\alpha$ in all our experiments.}

\begin{table}[htbp] 
\caption{The mAP, ACE, ECE, and MCE of our DCLR method on the SSGRL backbone with different $\alpha$ settings.}
\centering
\resizebox{0.7\linewidth}{!}{
\begin{tabular}{c|cccc}
\hline
 & mAP & ACE & ECE & MCE \\ \hline\hline
0.02 & 84.1 & 1.697 & 1.863 & 85.927 \\ 
0.03 & 84.1 & 1.602 & 1.795 & 80.687 \\
0.05 & 84.0	& 1.459	& 1.503 & 59.018 \\
0.1  & 83.9	& 1.923 & 2.072 & 89.258 \\
\hline
\end{tabular}}
\label{tab:alpha}
\end{table}

\subsubsection{Analysis of ILCRA}

\begin{table}[htbp]
\caption{The mAP, ACE, ECE, and MCE of the LS baseline, Ours ILCRA w/o $\mathcal{L}_c$, and Ours ILCRA.}
\centering
\resizebox{0.8\linewidth}{!}{
\begin{tabular}{c|cccc}
\hline

 & mAP & ACE & ECE & MCE \\ \hline\hline
LS  & 83.9	& 2.104 &	1.741 &	69.516 \\
Ours ILCAR w/o $\mathcal{L}_{acl}$ & 83.9 & 2.124 & 1.748 & 70.846\\
Ours ILCAR w/o $\mathcal{L}_c$ & 83.9 & 1.831 & 1.658 & 66.291\\ 
Ours ILCAR & 83.9 & 1.789  & 1.620 & 64.028\\ 
\hline
\end{tabular}}
\label{tab:ilcra}
\end{table}

ILCAR is designed to learn instance-level correlations, aiding in the creation of soft label vectors that can model semantic confusion. To evaluate its effectiveness, we conducted experiments solely utilizing ILCAR (namely Ours ILCRA) and compared it with the Label Smoothing (LS) baseline. The experimental results, illustrated in Table~\ref{tab:ilcra}, reveal that even the exclusive use of ILCRA can enhance calibration performance. In comparison to the LS baseline, ILCRA reduces the ACE, ECE, and MCE by 0.315, 0.121, and 5.488, respectively.

The auxiliary classification and contrastive losses can help to better learn category information and their correlations, and thus facilitate the construction of more effective regularization. Here, we also verify their contributions by comparing our model, Ours ILCRA, with two modified baselines that {omit} these two losses (denoted as Ours ILCRA w/o $\mathcal{L}_acl$ and Ours ILCRA w/o $\mathcal{L}_c$). As shown in Table \ref{tab:ilcra}, the exclusion of either loss leads to a notable decrease in ACE, ECE, and MCE. We observed that removing the auxiliary classification loss resulted in even poorer performance compared to the LS baseline. This could be attributed to the model's inability to guarantee that the learned features encompass information specific to the corresponding category, thereby hindering its ability to accurately learn correlations for constructing reasonable pseudo labels. Conversely, our model, even in the absence of the contrastive loss, surpassed the LS baseline, achieving reductions in ACE and ECE by 0.273 and 0.083, respectively. This discovery suggests that the inherent similarities present in category-specific features, which capture semantic category correlations, can enhance calibration outcomes without requiring retraining.

\subsubsection{Analysis of PLCRA}
\begin{table}[htbp]
\caption{The mAP, ACE, ECE, and MCE of the LS baseline, Ours PLCRA w/o $\mathcal{L}_c$, and Ours PLCRA.}
\centering
\resizebox{0.8\linewidth}{!}{
\begin{tabular}{c|cccc}
\hline
 & mAP & ACE & ECE & MCE \\ \hline\hline
LS  & 83.9	& 2.104 &	1.741 &	69.516 \\ 
Ours PLCRA w/o $\mathcal{L}_{acl}$ & 84.0 & 2.146 & 1.790 & 70.571\\
Ours PLCRA w/o $\mathcal{L}_c$ & 84.0 & 1.895 & 1.697 & 69.081\\ 
Ours PLCRA & 84.0 & 1.801 & 1.593 & 67.293\\ 
\hline
\end{tabular}}
\label{tab:plcra}
\end{table}

PLCRA aims to learn more robust prototype-level correlations to construct soft label vectors. Here, we also evaluate its contribution by merely using PLCRA (namely Ours PLCRA) and comparing it with the LS baseline. As illustrated in Table \ref{tab:plcra}, it also obtains better calibration performance, with the reductions of the ACE, ECE, and MCE by 0.303, 0.148, and 2.223, respectively. 

The contrastive loss can help to learn more compact representations and thus obtain more robust prototype-instance similarities. In this part, we design a new baseline without the contrastive loss (namely Ours PLCRA w/o $\mathcal{L}_c$) and compare it with Ours PLCRA. As illustrated in Table \ref{tab:plcra}, omitting the contrastive loss leads to {a} notable performance drop in all three metrics. Notably, Ours PLCRA w/o $\mathcal{L}_c$ can also lead to overall better calibration performance compared with the LS baseline, again demonstrating the existence of inherent category correlations without retraining. Similarly, removing the auxiliary classification loss (namely Ours PLCRA w/o $\mathcal{L}_{acl}$) leads to worse performance than both our PLCRA and the LS baseline.

{
$K$, the number of prototype-level features, is also a key important factor that affects the performance. 
To investigate its contribution, we conduct experiments with different $K$s on the SSGRL with our DCLR. The $K$ is set to 5, 10, and 20, respectively. The results are shown in Table~\ref{tab:k}. We can see that although all the models with different $K$ can achieve competitive results, a suitable $K$ is still important for confidence calibration since different images have different visual variations and a suitable $K$ can achieve a better trade-off to represent them. According to the experiment results, we choose 10 as the default value of $K$.  
}

\begin{table}[htbp] 
\caption{The mAP, ACE, ECE, and MCE of our DCLR method on the SSGRL backbone with different $K$ prototype-level features.}
\centering
\resizebox{0.7\linewidth}{!}{
\begin{tabular}{c|cccc}
\hline
 $K$ & mAP & ACE & ECE & MCE \\ \hline\hline
5 & 83.9 & 1.582 & 1.736 & 70.689 \\ 
10 & 84.0	& 1.459	& 1.503 & 59.018 \\
20 & 83.8	& 1.768	& 2.052 & 92.741 \\
\hline
\end{tabular}}
\label{tab:k}
\end{table}

\subsection{{Qualitative Comparison}}
\begin{figure*}[t]
\centering
\subfigure[] {\includegraphics[width=.49\textwidth]{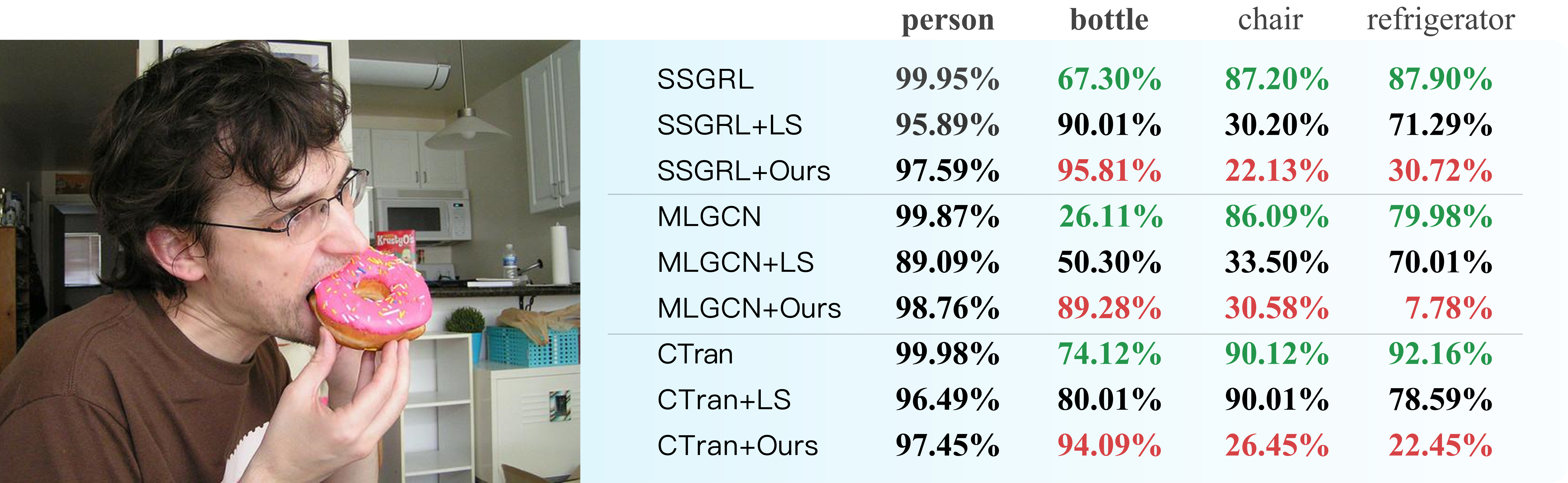}}
\subfigure[] {\includegraphics[width=.49\textwidth]{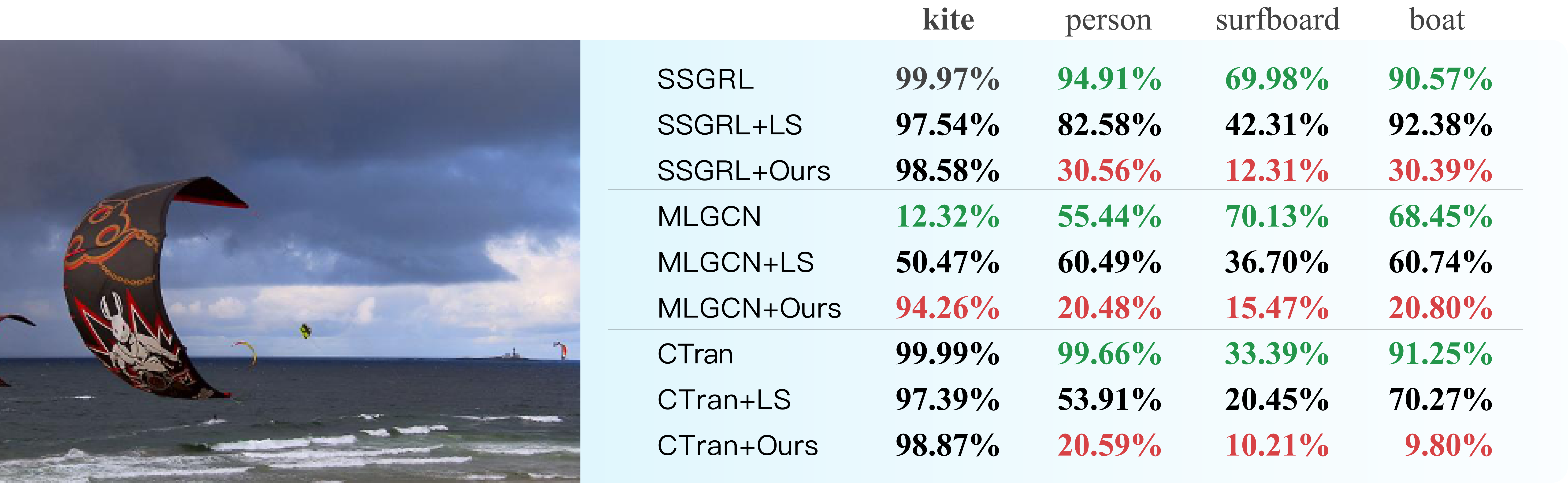}}
\subfigure[] {\includegraphics[width=.49\textwidth]{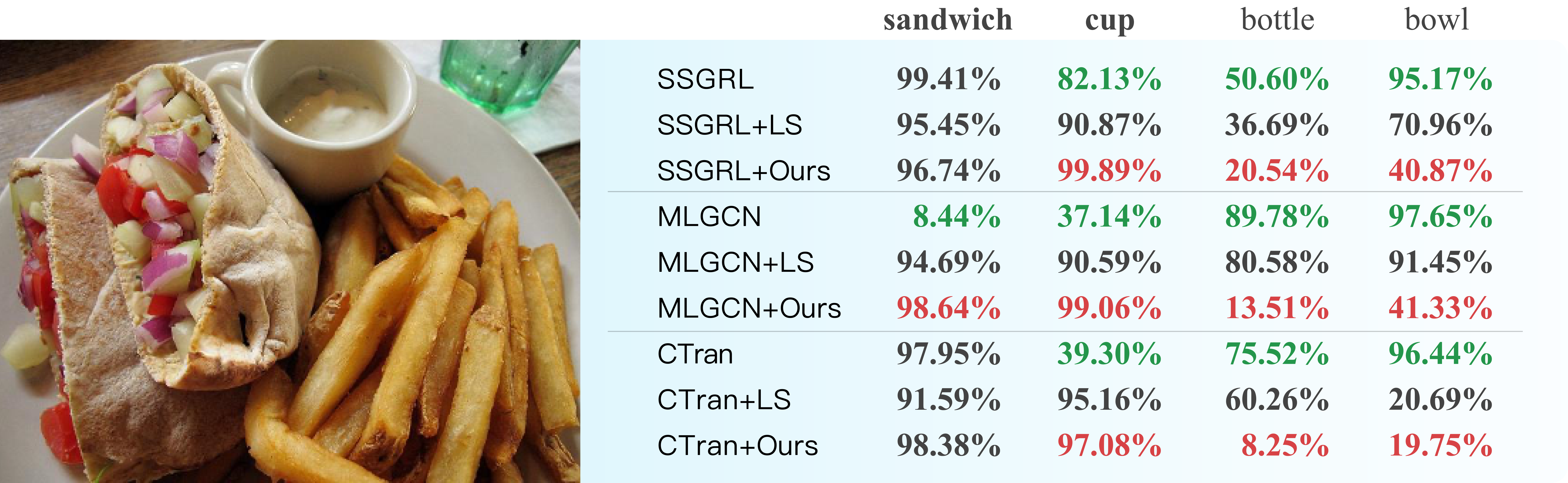}}
\subfigure[] {\includegraphics[width=.49\textwidth]{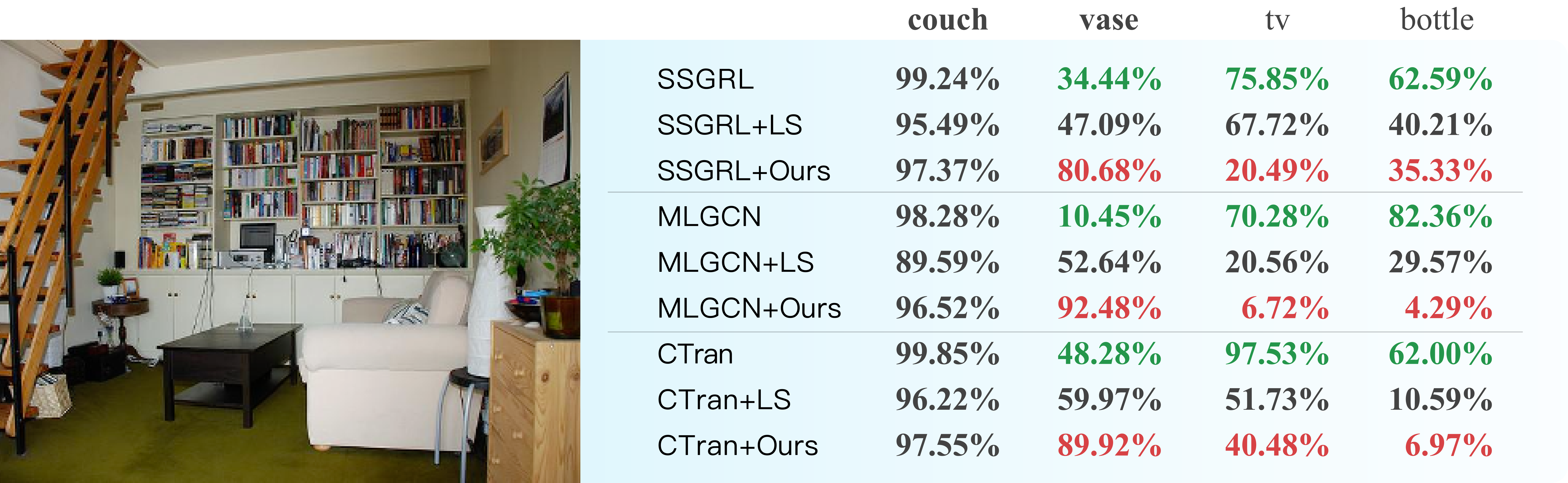}}
\caption{The visualization examples from the MS-COCO dataset. These examples can illustrate how our DCLR calibrates the confidence for better predictions.}
\label{fig:visual}
\end{figure*}

{To exhibit how our DCLR calibrates the predicted confidences, we provide some visualization examples from the MS-COCO dataset with predicted class confidences on different MLR models and different calibration methods in Fig.~\ref{fig:visual}. From the visualization examples (a), (b), (c), and (d), we can conclude that our DCLR can perform better confidence score prediction than the original MLR models and their enhanced versions with label smoothing (LS) since our DCLR can predict fairly high confidence score for correct classes while predicting lower confidence score for error classes, showing clearer discriminative boundaries. For example, in example (a), SSGRL, MLGCN, and CTran fail to predict `bottle` due to lower predicted confidence scores. Calibrating with label smoothing (LS), only SSGRL can predict `bottle` correctly while both MLGCN and CTran still suffer from lower predicted confidence scores of `bottle`. Benefiting from our proposed DCLR, all three models SSGRL, MLGCN, and CTran can predict `bottle` correctly with higher confidence scores than those wrong classes. Similar observations and results can be concluded from other examples. These results further show the effectiveness of our DCLR on multi-label confidence calibration since it can calibrate the predicted confidence better.   
}

\section{Conclusion}
In summary, this work tackles the prevalent overconfidence dilemma in multi-label scenarios, a challenge that has been largely neglected by existing single-label calibration algorithms. We introduce the innovative Dynamic Correlation Learning and Regularization (DCLR) algorithm, which harnesses multi-grained semantic correlations to model the semantic confusion characteristic of multi-label images, thereby ensuring well-calibrated confidence scores. DCLR utilizes dynamic instance-level and prototype-level similarities to construct adaptive label vectors that enable more effective regularization. Additionally, we have established a comprehensive benchmark to address the previously existing void in MLCC evaluations. This benchmark re-implements and adapts several state-of-the-art confidence calibration algorithms to the MLCC task. It also provides a performance comparison of these algorithms, alongside our proposed DCLR, integrated within three seminal MLR models across two widely-used datasets. 
{In the future, MLCC have two particularly important and meaningful research aspects. First, we currently consider only pairwise correlations between categories, a form of local correlations. It would be valuable and significant to consider all categories and introduce correlations from a holistic perspective, to better learn correlation information and thus achieve improved regularization. Second, there are many scenarios where labels are limited, such as multi-label recognition with partial labels and few-shot multi-label recognition. MLCC could provide more reliable labels to help effectively retrieve labels, which could enhance the performance of these practical tasks significantly.}

\bibliographystyle{IEEEtran}
\bibliography{reference}

\begin{IEEEbiography}[{\includegraphics[width=1in,height=1.25in,clip,keepaspectratio]{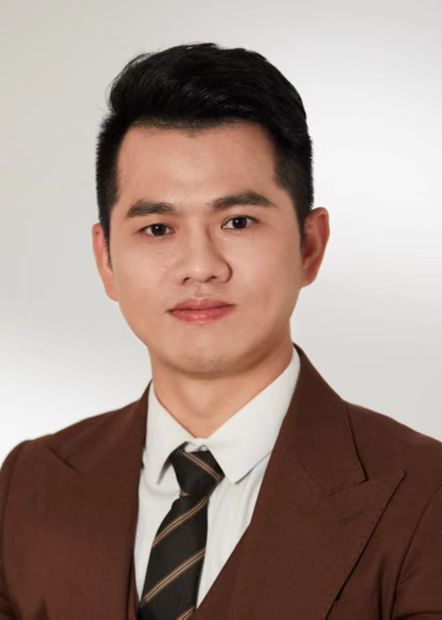}}]{Tianshui Chen} received a Ph.D. degree in computer science at the School of Data and Computer Science Sun Yat-sen University, Guangzhou, China, in 2018. Prior to earning his Ph.D, he received a B.E. degree from the School of Information and Science Technology in 2013. He is currently an associate professor at the Guangdong University of Technology. His current research interests include computer vision and machine learning. He has authored and co-authored more than 40 papers published in top-tier academic journals and conferences, including T-PAMI, T-NNLS, T-IP, T-MM, CVPR, ICCV, AAAI, IJCAI, ACM MM, etc. He has served as a reviewer for numerous academic journals and conferences. He was the recipient of the Best Paper Diamond Award at IEEE ICME 2017. \end{IEEEbiography}

\begin{IEEEbiography}[{\includegraphics[width=1in,height=1.25in,clip,keepaspectratio]{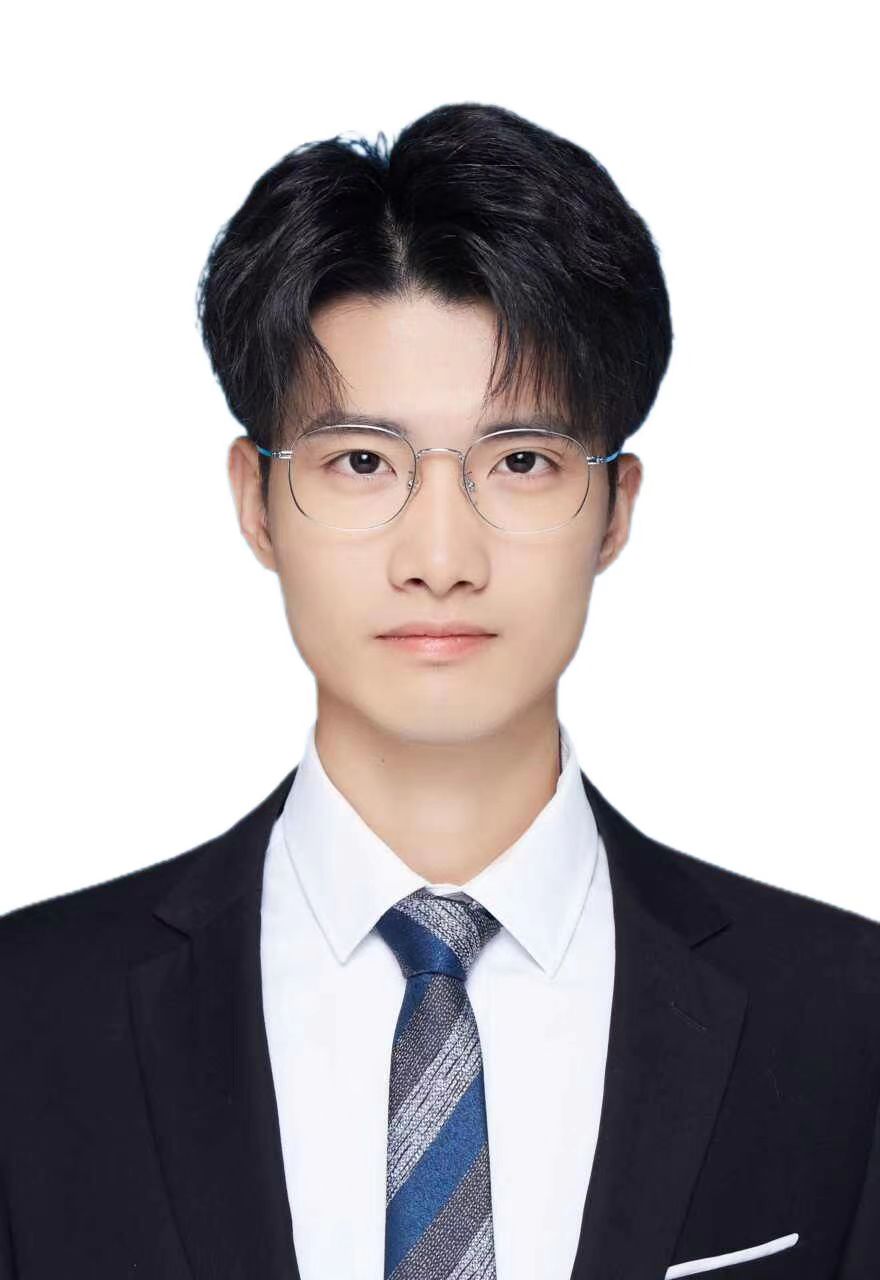}}]{Weihang Wang} is an undergraduate student majoring in Computer Science at Guangdong University of Technology. He will pursue a Ph.D. degree in Computer Science at Fudan University in 2024. His research interests lie in machine learning, particularly in computer vision and large language models.\end{IEEEbiography}

\begin{IEEEbiography}[{\includegraphics[width=1in,height=1.25in,clip,keepaspectratio]{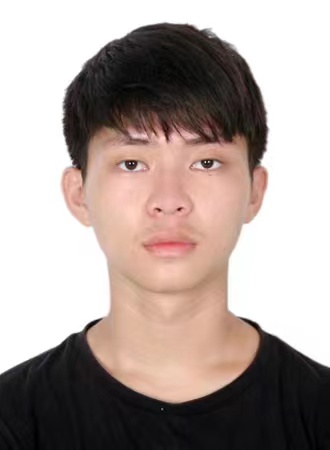}}]{Tao Pu} received a B.E. degree from the School of Computer Science and Engineering, Sun Yat-sen University, Guangzhou, China, in 2020, where he is currently pursuing a Ph.D. degree in computer science. He has authored and coauthored approximately 10 papers published in top-tier academic journals and conferences, including T-PAMI, AAAI, ACM MM, etc.\end{IEEEbiography}

\begin{IEEEbiography}[{\includegraphics[width=1in,height=1.25in,clip,keepaspectratio]{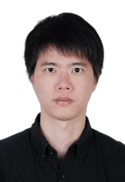}}]{Jinghui Qin} is a lecturer at the Guangdong University of Technology, Guangzhou, China. He received his B.S. and MA.Eng degrees from the School of Software, Sun Yat-Sen University, Guangzhou, China, in 2012 and 2014, respectively. He also received his Ph.D degree from the School of Data and Computer Science, Sun Yat-Sen University, Guangzhou, China, in 2020. He was a Post-Doctoral Fellow at the Sun Yat-sen University, Guangzhou, China, during 2020 and 2022. He has been serving as a reviewer of IEEE Trans. Neural Networks and Learning Systems, IEEE Trans. Broadcasting, Neurocomputing, Computer Vision and Image Understanding, Bioinformatics, etc. His research interest includes natural language processing, machine learning, and computer vision.\end{IEEEbiography}

\begin{IEEEbiography}[{\includegraphics[width=1in,height=1.25in,clip,keepaspectratio]{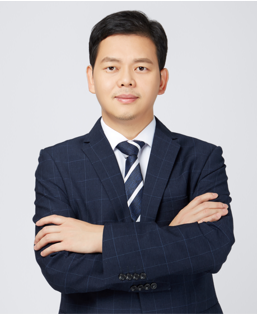}}]{Zhijing Yang} received the B.S and Ph.D. degrees from the Mathematics and Computing Science, Sun Yat-sen University, Guangzhou China, in 2003 and 2008, respectively. He was a Visiting Research Scholar in the School of Computing, Informatics and Media, University of Bradford, U.K, between July-Dec, 2009, and a Research Fellow in the School of Engineering, University of Lincoln, U.K, between Jan. 2011 to Jan. 2013. He is currently a Professor and Vice Dean at the School of Information Engineering, Guangdong University of Technology, China. He has published over 80 peer-reviewed journal and conference papers, including IEEE T-CSVT, T-MM, T-GRS, PR, etc. His research interests include machine learning and pattern recognition.
\end{IEEEbiography}

\begin{IEEEbiography}[{\includegraphics[width=1in,height=1.25in,clip,keepaspectratio]{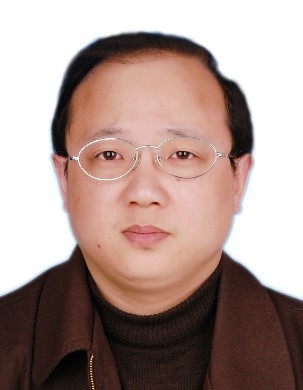}}]{Jie Liu} is now a professor and doctoral supervisor at School of Information Science, North China University of Technology, Beijing, P. R. China. He had worked in Capital Normal University from 1999 to 2021, He is currently a part-time doctoral advisor and master's supervisor of Capital Normal University. He received PH.D degree in Computer Application Technology at Beijing Institute of Technology in 2009. He is a deputy director of China language intelligence research center, a director of China Artificial Intelligence Society and the secretary-general of Language Intelligence Special Committee of China Artificial Intelligence Society. He won the first prize of Wu Wenjun's AI Science and Technology Progress in 2019, and the second prize of Beijing Science and Technology Progress in 2017, and Second Prize of Beijing Invention Patent Award in 2023. He is the project director of Science and Technology Innovation 2030 Major Project of the Ministry of Science and Technology of China. His main research interests are Natural Language Processing and Artificial Intelligence.\end{IEEEbiography}

\begin{IEEEbiography}[{\includegraphics[width=1in,clip]{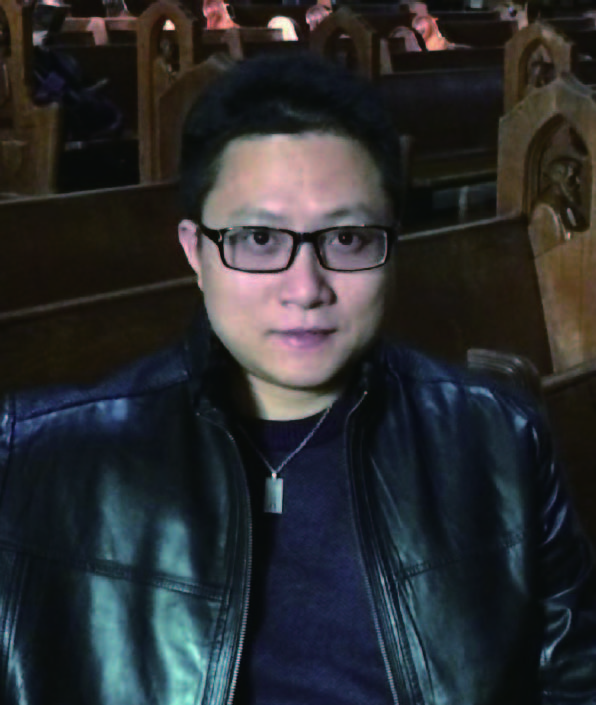}}]{Liang Lin} (Fellow, IEEE) is a full professor at Sun Yat-sen University. From 2008 to 2010, he was a postdoctoral fellow at the University of California, Los Angeles. From 2016--2018, he led the SenseTime R\&D teams to develop cutting-edge and deliverable solutions for computer vision, data analysis and mining, and intelligent robotic systems. He has authored and co-authored more than 100 papers in top-tier academic journals and conferences (e.g., 15 papers in TPAMI and IJCV and 60+ papers in CVPR, ICCV, NIPS, and IJCAI). He has served as an associate editor of IEEE Trans. Human-Machine Systems, The Visual Computer, and Neurocomputing and as an area/session chair for numerous conferences, such as CVPR, ICME, ACCV, and ICMR. He was the recipient of the Annual Best Paper Award by Pattern Recognition (Elsevier) in 2018, the Best Paper Diamond Award at IEEE ICME 2017, the Best Paper Runner-Up Award at ACM NPAR 2010, Google Faculty Award in 2012, the Best Student Paper Award at IEEE ICME 2014, and the Hong Kong Scholars Award in 2014. He is a Fellow of IEEE, IAPR, and IET. \end{IEEEbiography}

\vfill

\end{document}